\PassOptionsToPackage{dvipsnames}{xcolor}
\documentclass[11pt,a4paper]{article}

\usepackage[hyperref]{acl2019}
\usepackage{times}
\usepackage{latexsym}
\usepackage{url}
\usepackage[utf8]{inputenc}
\usepackage{latexsym}
\usepackage{amsmath}
\usepackage{amssymb}
\usepackage{amsfonts}       % blackboard math symbols
\usepackage{nicefrac}       % compact symbols for 1/2, etc.
\usepackage{microtype}      % microtypography
\usepackage{pgf}
\usepackage{apalike}
\usepackage{graphicx}
\usepackage{paralist}
\usepackage{tikz}
\usepackage{xspace}
\usepackage{subcaption}
\usepackage{physics}
\usetikzlibrary{bayesnet}
\usepackage{comment}
\usepackage[colorinlistoftodos,prependcaption,textsize=tiny]{todonotes}
\usepackage{bm}
\usepackage{booktabs}
\usepackage{adjustbox}
\usepackage{caption,subcaption}
\usepackage{float}
\restylefloat{table}
\usepackage{multirow}
\usepackage{cleveref}

\aclfinalcopy % Uncomment this line for the final submission
\definecolor{darkblue}{rgb}{0, 0, 0.5}
\hypersetup{colorlinks=true,citecolor=darkblue, linkcolor=darkblue, urlcolor=darkblue}
\newcommand{\red}[1]{\textcolor{Red}{#1}}

\newcommand{\bluebf}[1]{\textbf{\textcolor{Blue}{#1}}}

\newcommand{\green}[1]{\textcolor{OliveGreen}{#1}}

\newcommand{\white}[1]{\textcolor{White}{#1}}

\newcommand{\orangebf}[1]{\textbf{\textcolor{Orange}{#1}}}
\DeclareMathOperator{\avg}{avg}
\DeclareMathOperator{\KL}{KL}

\DeclareMathOperator{\softmax}{softmax}

\DeclareMathOperator{\softplus}{softplus}
\DeclareMathOperator{\diag}{diag}
\DeclareMathOperator{\affine}{affine}

\DeclareMathOperator{\emb}{emb}

\DeclareMathOperator{\detach}{detach}
\DeclareMathOperator{\ReLU}{ReLU}
\DeclareMathOperator{\attention}{attention}
\DeclareMathOperator{\BiLSTM}{BiLSTM}
\DeclareMathOperator{\LSTM}{LSTM}
\DeclareMathOperator{\MLP}{MLP}

\DeclareMathOperator{\Cat}{Cat}

\newcommand{\cond}{VMMT$_{\text{C}}$\xspace}
\newcommand{\uncond}{VMMT$_{\text{F}}$\xspace}

\newcommand{\NDist}[2]{\mathcal{N}\left(#1,#2\right)}
\DeclareMathOperator{\KullbackLeibler}{KL}

\title{Latent Variable Model for Multi-modal Translation}
\author{Iacer Calixto \\\And
  Miguel Rios \\
  ILLC \\
  The University of Amsterdam \\
  {\tt \{iacer.calixto,m.riosgaona,w.aziz\}@uva.nl} \\\And
  Wilker Aziz \\
  }

\begin{document}
\maketitle

\begin{abstract}
In this work, we propose to model the interaction between visual and textual features for multi-modal neural machine translation (MMT) through a latent variable model.
This latent variable can be seen as a multi-modal stochastic embedding of an image and its description in a foreign language.
It is used in a target-language decoder and also to predict image features.
Importantly, our model formulation utilises visual and textual inputs during training but does not require that images be available at test time.
We show that our latent variable MMT formulation improves considerably over strong baselines, including a multi-task learning approach~\citep{ElliottKadar2017} and a conditional variational auto-encoder approach~\citep{Toyamaetal2016}.
Finally, we show improvements due to
\textit{(i)} predicting image features in addition to only conditioning on them,
\textit{(ii)} imposing a constraint on the minimum amount of information encoded in the latent variable, and
\textit{(iii)} by training on additional target-language image descriptions (i.e. synthetic data).
\end{abstract}

\section{Introduction}

Multi-modal machine translation (MMT) is an exciting novel take on machine translation (MT) where we are interested in learning to translate sentences \textit{in the presence of visual input} (mostly images).
In the last three years there have been shared tasks~\citep{Speciaetal2016,Elliottetal2017,Barraultetal2018} where many research groups proposed different techniques to integrate images into MT, e.g.~\citet{Caglayanetal2017,LibovickyHelcl2017}.

Most MMT models expand neural machine translation (NMT) architectures \citep{Sutskever+2014:SSNN,BahdanauChoBengio2015} to additionally condition on an image in order to compute the likelihood of a translation in context. 
This gives the model a chance to exploit correlations in visual and language data, but also means that images must be available at test time.
An exception to this rule is the work of \citet{Toyamaetal2016} who exploit the framework of conditional variational auto-encoders (CVAEs)~\citep{CVAE} to decouple the encoder used for posterior inference at training time from the encoder used for generation at test time.
Rather than conditioning on image features, the model of \citet{ElliottKadar2017} learns to rank image features using language data in a multi-task learning (MTL) framework, therefore sharing parameters between a translation (generative) and a sentence-image ranking model (discriminative).
This similarly exploits correlations between the two modalities
and has the advantage that images are also not necessary at test time.

In this work, we also aim at translating without images at test time, yet learning a visually grounded translation model.
To that end, we resort to probabilistic modelling instead of multi-task learning and estimate a \textit{joint distribution} over translations and images.
In a nutshell, we propose to model the interaction between visual and textual features through a latent variable. This latent variable can be seen as a stochastic embedding which is used in the target-language decoder, as well as to \emph{predict} image features. 
Our experiments show that this joint formulation improves over an MTL approach \citep{ElliottKadar2017}, which does model both modalities but not jointly, and over the CVAE of \citet{Toyamaetal2016}, which uses image features to condition an inference network but crucially does not model the images.

The main contributions of this paper are:\footnote{Code and pre-trained models will be released soon.}
\begin{itemize}
  \item we propose a novel multi-modal NMT model that incorporates image features through \emph{latent variables} in a \emph{deep generative model}.
  \item our latent variable MMT formulation improves considerably over strong baselines, and compares favourably to the state-of-the-art.
  \item we exploit correlations between both modalities at training time through a joint generative approach and do not require images at prediction time.
\end{itemize}

The remainder of this paper is organised as follows.
In \S\ref{sec:models}, we describe our variational MMT models.
In \S\ref{sec:exp}, we introduce the data sets we used and report experiments and assess how our models compare to prior work.
In \S\ref{sec:related}, we position our approach with respect to the literature.
Finally, in \S\ref{sec:conclusions} we draw conclusions and provide avenues for future work.

\section{Variational Multi-modal NMT}\label{sec:models}

Similarly to standard NMT, in MMT we wish to translate a source sequence $x_1^m \triangleq \langle x_1, \cdots, x_m \rangle$ into a target sequence $y_1^n \triangleq \langle y_1, \cdots, y_n \rangle$.
The main difference is the presence of an image $v$ which illustrates the sentence pair $\langle x_1^m, y_1^n \rangle$. We do not model images directly, but instead an $2048$-dimensional vector of pre-activations of a ResNet-50's \texttt{pool5} layer~\cite{He2015}.

In our variational MMT models, image features are assumed to be generated by transforming a stochastic latent embedding $z$, which is 
also used to inform the RNN decoder in translating source sentences into a target language.

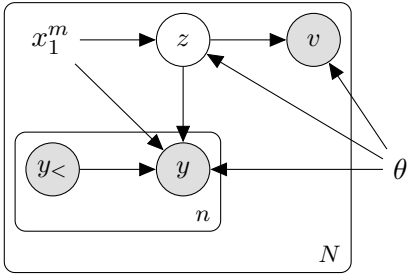
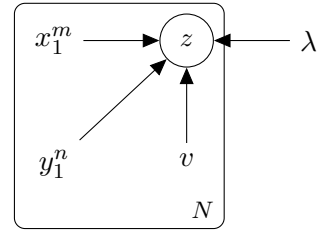
\begin{figure*}[ht!]
  \centering
  \begin{subfigure}[t]{0.45\linewidth}\centering
    \begin{tikzpicture}
    % Define nodes
    
    \node[obs]   (y) {$ y $};
    \node[obs, left=of y]   (yprev) {$ y_{<} $};
    \node[latent, above=of y]     (z) {$ z $};
    \node[left=of z]    (x) {$ x_1^m $};
    \node[obs, right=of z]   (v) {$ v $};
    \node[right=of y] (dummy) {};
    \node[right=of dummy] (theta) {$\theta$};

    % Connect nodes
    \edge{x}{z};
    \edge{x,z,yprev}{y};
    \edge{z}{v};
    \edge{theta}{y,z,v};

    % add plates
    \plate{y-sentence}{(y)(yprev)}{$n$};
    \plate{data}{(y-sentence)(x)(z)(v)}{$N$};

    \end{tikzpicture}
    \caption{\label{fig:vmmtc-gen}\cond: given the source text $x_1^m$, we model the joint likelihood of the translation $y_1^n$, the image (features) $v$, and a stochastic embedding $z$ sampled from a conditional latent Gaussian model. Note that the stochastic embedding is the sole responsible for assigning a probability to the observation $v$, and it helps assign a probability to the translation.}
  \end{subfigure}
  \hfill
  \begin{subfigure}[t]{0.45\linewidth}\centering
    \begin{tikzpicture}
    % Define nodes
    
    \node[]    (x) {$ x_1^m $};
    \node[below=of x]   (y) {$ y_1^n $};
    \node[latent, right=of x]     (z) {$ z $};
    
    \node[below=of z]   (v) {$ v $};    
    \node[right=of z] (theta) {$\lambda$};

    % Connect nodes
    \edge{x,y,v}{z};
    \edge{theta}{z};

    % add plates    
    \plate{data}{(y)(x)(z)(v)}{$N$};

    \end{tikzpicture}
    \caption{\label{fig:vmmtc-inf}Inference model for \cond: to approximate the true posterior we have access to both modalities (text $x_1^m, y_1^n$ and image $v$).}
  \end{subfigure}
  \caption{\label{fig:vmmt}Generative model of target text and image features (left), and inference model (right).}
\end{figure*}

\paragraph{Generative model} We propose a generative model of translation and image generation where both the image $v$ and the target sentence $y_1^n$ are independently generated given a common stochastic embedding $z$. 
The generative story is as follows.
We observe a source sentence $x_1^m$ and draw an embedding $z$ from a latent Gaussian model, 
\begin{equation}\label{eq:gaussian-latent-model}
\begin{aligned}
	Z|x_1^m &\sim \mathcal N(\boldsymbol\mu, \diag(\boldsymbol\sigma^2)) \\
    \boldsymbol\mu &= f_\mu(x_1^m; \theta) \\
    \boldsymbol\sigma &= f_\sigma(x_1^m; \theta) ~,
\end{aligned}
\end{equation}
where $f_\mu(\cdot)$ and $f_\sigma(\cdot)$ map from a source sentence to a vector of locations $\boldsymbol\mu \in \mathbb R^c$ and a vector of scales $\boldsymbol\sigma \in \mathbb R^c_{>0}$, respectively.
We then proceed to draw the image features from a Gaussian observation model,
\begin{equation}\label{eq:gaussian-obs-model}
\begin{aligned}
	V|z &\sim \mathcal N(\boldsymbol \nu, \varsigma^2 I) \\
    %V|z &\sim \mathcal N(\boldsymbol \nu, \diag(\boldsymbol \varsigma^2))  \\
    \boldsymbol \nu &= f_\nu(z; \theta) ~ , %\\
    %\boldsymbol \varsigma &= f_\varsigma(z; \theta) ~ ,
\end{aligned}
\end{equation}
where $f_\nu(\cdot)$ maps from $z$ to a vector of locations $\boldsymbol \nu \in \mathbb R^o$, and $\varsigma \in \mathbb R_{>0}$ is a hyperparameter of the model (we use $1$).
Conditioned on $z$ and on the source sentence $x_1^m$, and independently of $v$, we generate a translation by drawing each target word in context from a Categorical observation model,
\begin{equation}\label{eq:cat-obs-model}
\begin{aligned}
Y_j| x_1^m, z, y_{<j} &\sim \Cat(\boldsymbol \pi_j) \\
\boldsymbol \pi_j &= f_\pi(x_1^m, y_{<j}, z; \theta) ~, 
%\mathbf h_j &= \rnn_\theta(\mathbf h_{j-1}, \emb_\theta(y_{j-1}), \bm{z})
\end{aligned}
\end{equation}
where $f_\pi(\cdot)$ maps $z$, $x_1^m$, and a prefix translation $y_{<j}$ to the parameters $\boldsymbol \pi_j$ of a categorical distribution over the target vocabulary.
Functions $f_\mu(\cdot)$, $f_\sigma(\cdot)$, $f_\nu(\cdot)$, and $f_\pi(\cdot)$ are implemented as neural networks whose parameters are collectively denoted by $\theta$. In particular, implementing $f_\pi(\cdot)$ is as simple as augmenting a standard NMT architecture \citep{BahdanauChoBengio2015,Luongetal2015}, i.e. encoder-decoder with attention, with an additional input $z$ available at every time-step. All other functions are single-layer MLPs that transform the average encoder hidden state to the dimensionality of the corresponding Gaussian variable followed by an appropriate activation.\footnote{Locations have support on the entire real space, thus we use linear activations, scales must be strictly positive, thus we use a $\softplus$ activation.} 

Note that in effect we model a joint distribution 
\begin{equation}
\begin{aligned}
& p_\theta(y_1^n, v, z|x_1^m) =\\
&\quad p_\theta(z|x_1^m) p_\theta(v|z) P_\theta(y_1^n|x_1^m, z)
\end{aligned}
\end{equation}
consisting of three components which we parameterise directly.
As there are no observations for $z$, we cannot estimate these components directly.
We must instead marginalise $z$ out, which yields the marginal
\begin{equation}\label{eq:marginal}
\begin{aligned}
& P_\theta(y_1^n, v|x_1^m) = \\
&\int p_\theta(z|x_1^m) p_\theta(v|z) P_\theta(y_1^n|x_1^m, z) \dd z ~.
\end{aligned}
\end{equation}
An important statistical consideration about this model is that even though $y_1^n$ and $v$ are conditionally independent given $z$, they are \emph{marginally dependent}. 
This means that we have designed a data generating process where our observations $y_1^n, v|x_1^m$ are not assumed to have been independently produced.\footnote{This is an aspect of the model we aim to explore more explicitly in the near future.}
This is in direct contrast with multi-task learning or joint modelling without latent variables---for an extended discussion see \citep[\S~3]{eikema2018auto}.

Finally, Figure~\ref{fig:vmmt} (left) is a graphical depiction of the generative model: shaded circles denote observed random variables, unshaded circles indicate latent random variables, deterministic quantities are not circled; the internal plate indicates iteration over time-steps, the external plate indicates iteration over the training data. Note that deterministic parameters $\theta$ are global to all training instances, while stochastic embeddings $z$ are local to each tuple $\langle x_1^m, y_1^n, v \rangle$.

\paragraph{Inference} Parameter estimation for our model is challenging due to the intractability of the marginal likelihood function (\ref{eq:marginal}).
We can however employ variational inference (VI) \cite{Jordan+1999:VI}, in particular amortised VI \citep{Kingma+2014:VAE,Rezende+14:DGM}, and estimate parameters to maximise a lowerbound
\begin{equation}\label{eq:ELBO}%\label{eq:loss_function}
\begin{aligned}
%\log P_\theta(y_1^n,v|x_1^m) \ge \\
\mathbb E_{q_\lambda(z|x_1^m,y_1^n, v)}\left [\log  p_\theta(v|z) + \log P_\theta(y_1^n|x_1^m, z)  \right] \\
- \KL(q_\lambda(z|x_1^m, y_1^n, v)||p_\theta(z|x_1^m)) 
\end{aligned}
\end{equation}
 on the log-likelihood function. 
This evidence lowerbound (ELBO) is expressed in terms of an  \emph{inference model} $q_\lambda(z|x_1^m, y_1^n, v)$ which we design having tractability in mind. In particular, our \emph{approximate posterior} is a Gaussian distribution 
\begin{equation}%\label{eq:ELBO}
\begin{aligned}
q_\lambda(z|x_1^m, y_1^n, v) &= \mathcal N(z|\mathbf u, \diag(\mathbf s^2))\\
\mathbf u &= g_u(x_1^m, y_1^n, v; \lambda) \\
\mathbf s &= g_s(x_1^m, y_1^n, v; \lambda) \\
\end{aligned}
\end{equation}
parametrised by an \emph{inference network}, that is, an independently parameterised neural network (whose parameters we denote collectively by $\lambda$) which maps from observations, in our case a sentence pair and an image, to a variational location $\mathbf u \in \mathbb R^c$ and a variational scale $\mathbf s \in \mathbb R^c_{>0}$. 
Figure \ref{fig:vmmt} (right) is a graphical depiction of the inference model.
%\footnote{In general, an inference model can have access to all observations, $\langle x_1^m, y_1^n, v\rangle$ in our case. We restrict our model to conditioning on the source sentence alone, and we do so in order to enable the use of $q$ at test time when $y_1^n$ and $v$ are not known.}

Location-scale variables (e.g. Gaussians) can be reparametrised, i.e. we can obtain a latent sample via a deterministic transformation of the variational parameters and a sample from the standard Gaussian  distribution:
\begin{equation}
\begin{aligned}
z &= \mathbf u + \epsilon \odot \mathbf s  & \text{where } \epsilon \sim \mathcal N(0, I) ~.
\end{aligned}
\end{equation}
This reparametrisation enables backpropagation through stochastic units \citep{Kingma+2014:VAE,TitsiasEtAl2014doubly}.
In addition, for two Gaussians the $\KL$ term in the ELBO (\ref{eq:ELBO}) can be computed in closed form \citep[Appendix B]{Kingma+2014:VAE}.
 Altogether, we can obtain a reparameterised gradient estimate of the ELBO, we use a single sample estimate of the first term,  and count on stochastic gradient descent to attain a local optimum of (\ref{eq:ELBO}).

\paragraph{Architecture} 
All of our parametric functions are neural network architectures. 
In particular, $f_\pi$ is a standard sequence-to-sequence architecture with attention and a $\softmax$ output. We build upon OpenNMT \citep{opennmt}, which we modify slightly by providing $z$ as additional input to the target-language decoder at each time step.
Location layers $f_\mu$, $f_\nu$ and $g_u$, and scale layers $f_\sigma$ and $g_s$, are feed-forward networks with a single ReLU hidden layer. Furthermore, location layers have a linear output while scale layers have a $\softplus$ output.
For the generative model, $f_\mu$ and $f_\sigma$ transform the average source-language encoder hidden state. We let the inference model condition on source-language encodings without updating them, and we use a target-language bidirectional LSTM encoder in order to also condition on the complete target sentence. Then $g_u$ and $g_s$ transform a concatenation of the average source-language encoder hidden state, the average target-language bidirectional encoder hidden state, and the image features.

\paragraph{Fixed Gaussian prior} We have just presented our variational MMT model in its full generality---we refer to that model as \cond. 
However, keeping in mind that MMT datasets are rather small, it is desirable to simplify some of our model's  components. In particular, the estimated latent Gaussian model (\ref{eq:gaussian-latent-model}) can be replaced by a fixed standard Gaussian prior, i.e., $Z \sim \mathcal N(0, I)$---we refer to this model as \uncond.  Along with this change it is convenient to modify the inference model to condition on $x_1^m$ alone, which allow us to use the inference model for both training and prediction. Importantly this also sidesteps the need for a target-language bidirectional LSTM encoder, which leaves us a smaller set of inference parameters $\lambda$ to estimate. Interestingly, this model does not rely on features from $v$, instead only using it as learning signal through the objective in (\ref{eq:ELBO}), which is in direct contrast with the model of \citet{Toyamaetal2016}.

\section{Experiments}\label{sec:exp}

\begin{table*}[t!]
\centering
%\resizebox{\linewidth}{!} {
\begin{tabular}{lllllllll}
\toprule
{\bf Model} & \multicolumn{2}{c}{\bf BLEU4$\uparrow$} & \multicolumn{2}{c}{\bf METEOR$\uparrow$} & \multicolumn{2}{c}{\bf chrF$\uparrow$} & \multicolumn{2}{c}{\bf BEER$\uparrow$} \\
\midrule
NMT  &
	35.0 (0.4) && 54.9 (0.2) && \underline{61.0 (0.2)} && \underline{65.2 (0.1)} \\
Imagination &
	\underline{36.8 (0.8)} &&  55.8 (0.4) && -- && -- \\
Model G &
	36.5 && \underline{{\bf 56.0}} && -- && -- \\
    
\midrule

\uncond  &
	{\bf 37.6 (0.4)} & \green{$\uparrow 0.8$} &
    {\bf 56.0 (0.1)} & \green{$\uparrow 0.0$} &
    {\bf 62.1 (0.1)} & \green{$\uparrow 1.1$} &
    {\bf 66.6 (0.1)} & \green{$\uparrow 1.4$} \\
\cond  &
	37.4 (0.3) & \green{$\uparrow 0.6$} &
    55.8 (0.1) & \red{$\downarrow 0.2$} &
    62.0 (0.1) & \green{$\uparrow 1.0$} &
    66.5 (0.1) & \green{$\uparrow 1.3$} \\
\bottomrule
\end{tabular}
%}
\caption{Results of applying variational MMT models to translate the Multi30k test set. For each model, we report the mean and standard deviation over $4$ independent runs where models were selected using validation BLEU4 scores. Best mean baseline scores per metric are underlined and best overall results (i.e. means) are in bold. We highlight in green/red the improvement brought by our models compared to the best baseline mean score.}
\label{tab:main_results}
\end{table*}

Our encoder is a 2-layer $500$D bidirectional RNN with GRU, the latent embedding $z$, source, and target word embeddings are also $500$D each, and trained jointly with the model.
We use OpenNMT to implement all our models \citep{opennmt}.
All model parameters are initialised sampling from a uniform distribution $\mathcal{U}(-0.1,+0.1)$ and bias vectors are initialised to $\vec{0}$.

Visual features are obtained by feeding images to the pre-trained ResNet-50 and using the activations of the \texttt{pool5} layer~\cite{He2015}.
We apply dropout with a probability of $0.5$ in the encoder bidirectional RNN, the image features, the decoder RNN, and before emitting a target word.

All models are trained using the Adam optimiser~\cite{KingmaBa2014} with an initial learning rate of $0.002$ and minibatches of size 40, where each training instance consists of one English sentence, one German sentence and one image (MMT).
Models are trained for up to 40 epochs and we perform model selection based on BLEU4, and use the best performing model on the validation set to translate test data.
Moreover, we halt training if the model does not improve BLEU4 scores on the validation set for 10 epochs or more.
We report mean and standard deviation over $4$ independent runs for all models we trained ourselves (NMT, \uncond, \cond), and other baseline results are the ones reported in the authors' publications \citep{Toyamaetal2016,ElliottKadar2017}.

We preprocess our data by tokenizing, lowercasing, and converting words to subword tokens using a bilingual BPE model with $10$k merge operations~\cite{Sennrichetal2016}.
We quantitatively evaluate translation quality using case-insensitive and tokenized outputs in terms of BLEU$4$~\cite{Papinenietal2002}, METEOR~\cite{DenkowskiLavie2014}, chrF3~\cite{Popovic2015}, and BEER~\cite{Stanojevic2014}.
By using these, we hope to include word-level metrics which are traditionally used by the MT community (i.e. BLEU and METEOR), as well as more recent metrics which operate at the character level and that better correlate with human judgements of translation quality (i.e. chrF3 and BEER) \citep{Bojaretal2017}.

\subsection{Datasets}\label{sec:data}
The Flickr30k dataset~\cite{Youngetal2014} consists of images from Flickr and their English descriptions.
We use the \emph{translated Multi30k} (M30k$_\text{T}$) dataset~\citep{ElliottFrankSimaanSpecia2016}, i.e. an extension of Flickr30k where for each image one of its English descriptions was translated into German by a professional translator. Training, validation and test sets contain $29$k, $1014$ and $1$k images respectively, each accompanied by the original English sentence and its translation into German.

Since this dataset is very small, we also investigate the effect of including more in-domain data to train our models.
To that purpose, we use additional $145$K monolingual German descriptions released as part of the Multi30k dataset to the task of image description generation~\citep{ElliottFrankSimaanSpecia2016}.
We refer to this dataset as \emph{comparable Multi30k} (M30k$_\text{C}$).
Descriptions in the \emph{comparable Multi30k} were collected independently of existing English descriptions and describe the same $29$K images as in the M30k$_\text{T}$ dataset.

In order to obtain features for images, we use ResNet-50~\citep{He2015} pre-trained on ImageNet~\citep{ILSVRC15}.
We report experiments using
\texttt{pool5} features as our image features, i.e. $2048$-dimensional pre-activations of the last layer of the network.

\subsection{Baselines}\label{sec:baselines}
We compare our work against three different baselines.
The first one is a standard text-only sequence-to-sequence \textbf{NMT} model with attention 
\citep{Luongetal2015}, trained from scratch using hyperparameters described above.
The second baseline is the variational multi-modal MT model \textbf{Model G} proposed by \citet{Toyamaetal2016}, where global image features are used as additional input to condition an inference network.
Finally, a third baseline is the \textbf{Imagination} model of \citet{ElliottKadar2017}, a multi-task MMT model which uses a shared source-language encoder RNN and is trained in two tasks: to translate from English into German and on image-sentence ranking (English$\leftrightarrow$image).

\subsection{Translated Multi30k}
\label{sec:translated}

We now report on experiments conducted with models trained to translate from English into German using the \emph{translated Multi30k} data set (M30k$_\text{T}$).

In Table~\ref{tab:main_results}, we compare our variational MMT models---\cond for the general case with a conditional Gaussian latent model, and \uncond for the simpler case of a fixed Gaussian prior---to the three baselines described above.
The general trend is that both formulations of our VMMT improve with respect to all three baselines.
We note an improvement in BLEU and METEOR mean scores compared to the Imagination model \citep{ElliottKadar2017}, as well as reduced variance (though note this is based on only 4 independent runs in our case, and 3 independent runs of Imagination).
Both models \uncond and \cond outperform Model G according to BLEU and perform comparably according to METEOR, especially since results reported by \citep{Toyamaetal2016} are based on a single run.

Moreover, we also note that both our models outperform the text-only NMT baseline according to all four metrics, and by 1\%--1.4\% according chrF3 and BEER, both being metrics well-suited to measure the quality of translations into German and generated with subwords units.

Finally, one interesting finding is that the fixed-prior model \uncond performs slightly better than the conditional model \cond according to all four metrics studied.
We speculate this is due to \uncond's simpler parameterisation, after all, we have just about $29$k training instances to estimate two sets of parameters ($\theta$ and $\lambda$) and the more complex \cond requires an additional bidirectional LSTM encoder for the target text.

\subsection{Back-translated Comparable Multi30k}\label{sec:backtranslated}

Since the \emph{translated Multi30k} dataset is very small, %($\sim29$K triples of a sentence pair and an image that illustrates the sentences), 
we also investigate the effect of including more in-domain data to train our models.
For that purpose, we use additional $145$K monolingual German descriptions released as part of the \emph{comparable Multi30k} dataset (M30k$_\text{C}$). %, which were collected independently from the English descriptions but describe the same images in the original \emph{translated Multi30k}.
We train a text-only NMT model to translate from German into English using the original $29$K parallel sentences in the \emph{translated Multi30k} (without images), and apply this model to back-translate the $145$K German descriptions into English~\citep{Sennrichetal:2016:Backtranslation}.

\begin{figure}[t!]
  \centering
  \includegraphics[width=0.5\textwidth]{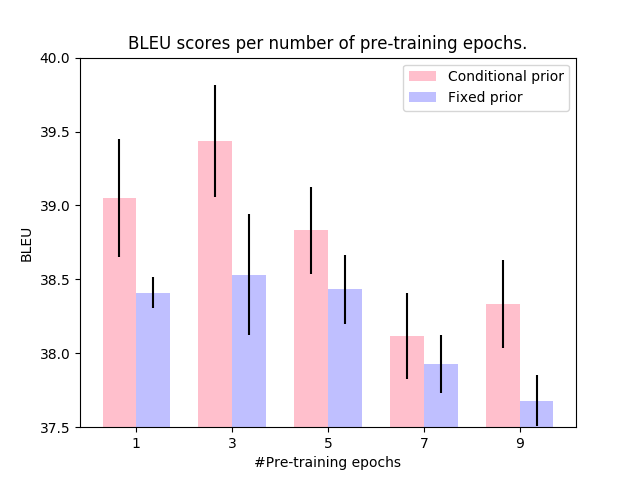}
  \caption{\label{fig:backtrans}Validation set BLEU scores per number of pre-trained epochs for models \cond and \uncond pre-trained using the \emph{comparable Multi30k} and \emph{translated Multi30k} data sets. The height of a bar represents the mean and the black vertical lines indicate $\pm 1$ std over 4 independent runs.}
\end{figure}

In this set of experiments, we explore how pre-training models NMT, \uncond and \cond using both the \emph{translated} and \emph{back-translated comparable Multi30k} affects results.
Models are pre-trained on mini-batches with a one-to-one ratio of \emph{translated} and \emph{back-translated} data.\footnote{One pre-training epoch corresponds to about $290$K examples, i.e. we up-sample the smaller \emph{translated Multi30k} data set to achieve the one-to-one ratio.}
All three models NMT, \uncond and \cond, are further fine-tuned on the translated Multi30k until convergence, and model selection using BLEU is only applied during fine-tuning and not at the pre-training stage.

In Figure~\ref{fig:backtrans}, we inspect for how many epochs should a model be pre-trained using the additional noisy back-translated descriptions, and
note that both \uncond and \cond reach best BLEU scores on the validation set when pre-trained for about 3 epochs.
As shown in Figure~\ref{fig:backtrans}, we note that when using additional noisy data \cond, which uses a conditional prior, performs considerably better than its counterpart \uncond, which has a fixed prior.
These results indicate that \cond makes better use of additional synthetic data than \uncond.
Some of the reasons that explain these results are
\textit{(i)} the conditional prior $p(z|x)$ can learn to be sensitive to whether $x$ is gold-standard or synthetic, whereas $p(z)$ cannot;
\textit{(ii)} in the conditional case the posterior approximation $q(z|x,y,v)$ can directly exploit different patterns arising from a gold-standard versus a synthetic $\langle x,y \rangle$ pair; and finally
\textit{(iii)} our synthetic data is made of \emph{target-language} gold-standard image descriptions, which help train the inference network's target-language BiLSTM encoder.

In Table~\ref{tab:backtrans_results}, we show results when applying \uncond and \cond to translate the Multi30k test set. Both models and the NMT baseline are pre-trained on the \emph{translated} and the \emph{back-translated comparable Multi30k} data sets, and are selected according to validation set BLEU scores.
For comparison, we also include results for Imagination~\citep{ElliottKadar2017} when trained on the \emph{translated Multi30k}, the WMT News Commentary English-German dataset ($240$K parallel sentence pairs) and the MSCOCO image description dataset ($414$K German descriptions of $83$K images, i.e. $5$ descriptions for each image).
In contrast, our models observe $29$K images (i.e. the same as the models evaluated in Section~\ref{sec:translated}) plus $145$K German descriptions only.\footnote{There are no additional images because the \emph{comparable Multi30k} consists of additional German descriptions for the same $29$K images already in the \emph{translated} Multi30k.}

\begin{table}[t!]
\centering
\resizebox{\linewidth}{!} {
\begin{tabular}{llll}
\toprule
{\bf Model} & {\bf BLEU4$\uparrow$} & {\bf METEOR$\uparrow$} & \bf \# train \\
&&& \bf sents. \\
\midrule
NMT  &
    37.7 (0.5) & 56.0 (0.3) & \multirow{3}{*}{$145$K}\\

\uncond  &
	{\bf 38.4 (0.6)} \white{-}\green{$\uparrow 0.7$} &
    56.0 (0.3) \white{-}\green{$\uparrow 0.0$} \\

\cond  &
	{\bf 38.4 (0.2)} \white{-}\green{$\uparrow 0.7$} &
    {\bf 56.3 (0.2)} \white{-}\green{$\uparrow 0.3$} \\

\midrule
Imagination &
	37.8 (0.7) &  57.1 (0.2) & $654$K \\
\bottomrule
\end{tabular}
}
\caption{Results for models pre-trained using the \emph{translated} and \emph{comparable Multi30k} to translate the Multi30k test set. We report the mean and standard deviation over $4$ independent runs.
Our best overall results are highlighted in bold, and we highlight in green/red the improvement/decrease brought by our models compared to the baseline mean score.
We additionally show results for the Imagination model trained on $4\times$ more data (as reported in the authors' paper).}
\label{tab:backtrans_results}
\end{table}

\subsection{Ablative experiments}
In our ablation we are interested in finding out to what extent the model makes use of the latent space, i.e. how important is the latent variable.

\paragraph{KL free bits}

A common issue when training latent variable models with a strong decoder is having the KL term in the loss collapse to the prior.
In practice, that would mean the model have virtually not used the latent variable $z$ to predict image features $v$, but mostly as a source of stochasticity in the decoder.
This can happen because the model has access to informative features from the source bi-LSTM encoder and need not learn a difficult mapping from observations to latent representations predictive of image features.

\begin{table}[t!]
\centering
%\resizebox{\linewidth}{!} {
\begin{tabular}{lll}
\toprule
{\bf Model} & {\bf Number of} & \bf BLEU4$\uparrow$ \\
			& {\bf free bits (KL)} &				\\
\midrule

\multirow{5}{*}{\uncond}	& 0 & 38.3 (0.2)  \\
									& 1 & 38.1 (0.3)  \\
									& 2 & {\bf 38.4 (0.4) }  \\
									& 4 & {\bf 38.4 (0.4) }  \\
									& 8 & 35.7 (3.1)  \\
\midrule
\multirow{5}{*}{\cond}	& 0 & 38.5 (0.2)  \\
									& 1 & 38.3 (0.3)  \\
									& 2 & 38.2 (0.2)  \\
									& 4 & 36.8 (2.6)  \\
									& 8 & {\bf 38.6 (0.2) } \\

\bottomrule
\end{tabular}
%}
\caption{Results of applying VMMT models trained with different numbers of free bits in the KL~\citep{Kingmaetal2016} to translate the Multi30k validation set.}
\label{tab:kl_free_bits}
\end{table}

\begin{table*}[t!]
  \centering
  \resizebox{\linewidth}{!} {
  \begin{tabular}{lllll}
    \textbf{Model} && \textbf{Example \#801} && \textbf{Example \#873} \\
    \midrule
    source & \multirow{10}{*}[-0.2em]{\includegraphics[width=0.23\linewidth]{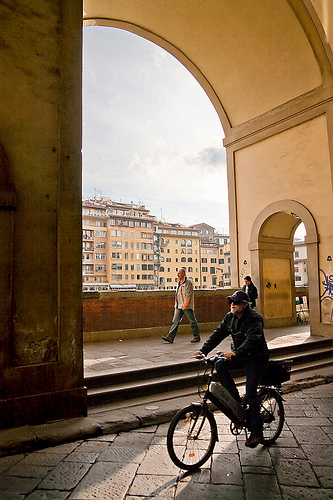}} & a man on a bycicle pedals through an \bluebf{archway} . &
    \multirow{10}{*}[-1.0em]{\includegraphics[width=0.35\linewidth]{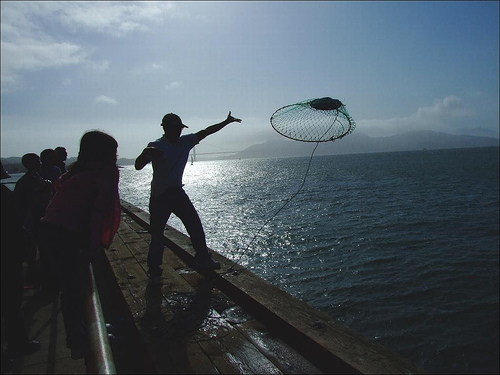}} & a man throws a fishing net into the \orangebf{bay} . \\
    reference &  & ein mann fährt auf einem fahrrad durch einen \bluebf{torbogen} . && ein mann wirft ein fischernetz in die \orangebf{bucht} . \\
    \cmidrule{1-1}\cmidrule{3-3}\cmidrule{5-5}
    && \textbf{M30k$_\text{T}$} && \textbf{M30k$_\text{T}$} \\
    \cmidrule{1-1}\cmidrule{3-3}\cmidrule{5-5}
    NMT && ein mann auf einem fahrrad fährt durch eine \bluebf{scheibe} . && ein mann wirft ein fischernetz in die \orangebf{luft} . \\
    \uncond &  &  ein mann auf einem fahrrad fährt durch einen \bluebf{torbogen} . && ein mann wirft ein fischernetz in die \orangebf{bucht} . \\
    \cond &  & ein mann auf einem fahrrad fährt durch einen \bluebf{bogen} . && ein mann wirft ein fischernetz in die \orangebf{bucht} . \\
    \cmidrule{1-1}\cmidrule{3-3}\cmidrule{5-5}
    && \textbf{M30k$_\text{T}$ + back-translated M30k$_\text{C}$} && \textbf{M30k$_\text{T}$ + back-translated M30k$_\text{C}$} \\
    \cmidrule{1-1}\cmidrule{3-3}\cmidrule{5-5}
    NMT && ein mann auf einem fahrrad fährt durch einen \bluebf{bogen} . && ein mann wirft ein fischernetz ins \orangebf{meer} . \\
    \uncond &  &  ein mann auf einem fahrrad fährt durch einen \bluebf{torbogen} . && ein mann wirft ein fischernetz in den \orangebf{wellen} . \\
    \cond &  & ein mann auf einem fahrrad fährt durch einen \bluebf{torbogen} . && ein mann wirft ein fischernetz in die \orangebf{bucht} . \\
  \end{tabular}
  }
  \caption{Translations for examples 801 and 873 of the M30k test set. In the first example, neither the NMT baseline (with or without back-translated data) nor model \cond (trained on limited data) could translate \bluebf{archway} correctly; the NMT baseline translates it as ``scheibe'' (disk) and ``bogen'' (bow), and \cond also incorrectly translates it as ``bogen'' (bow). However,  \cond translates without errors when trained on additional back-translated data, i.e. ``torbogen'' (archway).
  In the second example, the NMT baseline translates \orangebf{bay} as ``luft'' (air) or ``meer'' (sea), whereas \uncond translates it as ``bucht'' (bay) or ``wellen'' (waves) and \cond always as ``bucht'' (bay).
  %None of the different MT models could translate the example without errors, using additional back-translation data or otherwise. However, back-translated data have made errors arguably less damaging to the perceived quality of translations.}
  }
  \label{tab:example1}
\end{table*}

For that reason, we wish to measure how well can we train latent variable MMT models while ensuring that the KL term in the loss (Equation (\ref{eq:ELBO})) does not collapse to the prior.
We use the \emph{free bits} heuristic~\citep{Kingmaetal2016} to impose a constraint on the minimum amount of information the latent variable must encode therefore preventing it from collapsing.

In Table~\ref{tab:kl_free_bits}, we see the results of different models trained using different number of free bits in the KL component.
We note that including free bits improves translations slightly, but note that finding the number of free bits is not straightforward (i.e. it is a hyperparameter that needs searching).

\subsection{Discussion}
\label{sec:discussion}
In Table~\ref{tab:example1} we show how our different models translate two examples of the M30k test set.
In the first example (id\#801), training on additional back-translated data improves variational models but not the NMT baseline, whereas in the second example (id\#873) differences between baseline and variational models still persist even when training on additional back-translated data.

\section{Related work}
\label{sec:related}

Even though there has been growing interest in variational approaches to machine translation~\citep{Zhangetal2016,SchulzAzizCohn2018_ACL,ShahBarber2018_NIPS} and to tasks that integrate vision and language, e.g. image description generation~\citep{Yunchenetal2016_NIPS,Wangetal2017_NIPS}, relatively little attention has been dedicated to variational models for multi-modal translation.
This is partly due to the fact that multi-modal machine translation was only recently addressed by the MT community by means of a shared task~\cite{Speciaetal2016,Elliottetal2017,Barraultetal2018}.
Nevertheless, we now discuss relevant variational and deterministic multi-modal MT models in the literature.

\paragraph{Fully supervised MMT models.} All submissions to the three runs of the multi-modal MT shared tasks~\citep{Speciaetal2016,Elliottetal2017,Barraultetal2018} model conditional probabilities directly without latent variables.

Perhaps the first MMT model proposed prior to these shared tasks is that of \newcite{Hitschleretal2016}, who used image features to re-rank translations of image descriptions generated by a phrase-based statistical MT model (PBSMT) and reported significant improvements.
\citet{Shahetal2016} propose a similar model where image logits are used to re-rank the output of PBSMT.
Global image features, i.e. features computed over an entire image (such as \texttt{pool5} ResNet-50 features used in this work), have been directly used as ``tokens'' in the source sentence, to initialise encoder RNN hidden states, or as additional information used to initialise the decoder RNN states \citep{Huangetal2016,Libovickyetal2016,CalixtoLiu2017}.
On the other hand, spatial visual features, i.e. local features that encode different parts of the image separately in different vectors, have been used in doubly-attentive models where there is one attention mechanism over the source RNN hidden states and another one over the image features \citep{Caglayanetal2016,Calixtoetal2017}.

Finally, \citet{Caglayanetal2017} proposed to interact image features with target word embeddings, more specifically to perform an element-wise multiplication of the (projected) global image features and the target word embeddings before feeding the target word embeddings into their decoder GRU.
They reported significant improvements by using image features to gate target word embeddings and won the 2017 Multi-modal MT shared task \citep{Elliottetal2017}.

\paragraph{Multi-task MMT models.} Multi-task learning MMT models are easily applicable to translate sentences without images (at test time), which is an advantage over the above-mentioned models.

\newcite{Luongetal2016} proposed a multi-task approach where a model is trained using two tasks and a shared decoder: the main task is to translate from German into English and the secondary task is to generate English descriptions given an image.
They show improvements in the main translation task when also training for the secondary image description task.
Their model is large, i.e. a 4-layer encoder LSTM and a 4-layer decoder LSTM, and their best set up uses a ratio of $0.05$ image description generation training data samples in comparison to translation training data samples.
\citet{ElliottKadar2017} propose an MTL model trained to do translation (English$\rightarrow$German) and sentence-image ranking (English$\leftrightarrow$image), using a standard word cross-entropy and margin-based losses as its task objectives, respectively.
Their model uses the pre-trained GoogleNet v3 CNN \citep{Szegedyetal2016} to extract \texttt{pool5} features, and has a 1-layer source-language bidirectional GRU encoder and a 1-layer GRU decoder.

\paragraph{Variational MMT models.} \citet{Toyamaetal2016} proposed a variational MMT model that is likely the most similar model to the one we put forward in this work.
They build on the variational neural MT (VNMT) model of \citet{Zhangetal2016}, which is a conditional latent model where a Gaussian-distributed prior of $z$ is parameterised as a function of the the source sentence $x_1^m$, i.e. $p(z|x_1^m)$, and both $x_1^m$ and $z$ are used at each time step in an attentive decoder RNN, $P(y_j|x_1^m, z, y_{<j})$.

In \citet{Toyamaetal2016}, image features are used as input to the inference model $q_\lambda(z|x_1^m, y_1^n, v)$ that approximates the posterior over the latent variable, but otherwise are not modelled and not used in the generative network.
Differently from their work, we use image features in all our generative models, and propose modelling them as random observed outcomes while still being able to use our model to translate without images at test time.
In the conditional case, we further use image features for posterior inference. 
Additionally, we also investigate both conditional and fixed priors, i.e. $p(z|x_1^m)$ and $p(z)$, whereas their model is always conditional.
Interestingly, we found in our experiments that fixed-prior models perform slightly better than conditional ones under limited training data.

\citet{Toyamaetal2016} uses the pre-trained VGG19 CNN \citep{SimonyanZisserman2014} to extract FC7 features, and additionally experiment with using additional features from object detections obtained with the Fast RCNN network \citep{Girshick2015}.
One more difference between their work and ours is that we only use the ResNet-50 network to extract \texttt{pool5} features, and no additional pre-trained CNN nor object detections.

\section{Conclusions and Future work}\label{sec:conclusions}

We have proposed a latent variable model for multi-modal neural machine translation and have shown benefits from both modelling images and promoting use of latent space.
We also show that in the absence of enough data to train a more complex inference network a simple fixed prior suffices, whereas when more training data is available (even noisy data) a conditional prior is preferable. 
Importantly, our models compare favourably to the state-of-the-art.

In future work we will explore other generative models for multi-modal MT, as well as different ways to directly incorporate images into these models. We are also interested in modelling different views of the image, such as global vs. local image features, and also on modelling pixels directly (and using larger image collections).

%\section*{Acknowledgements}
%This work is supported by the Dutch Organisation for Scientific Research (NWO) VICI Grant nr. 277-89-002.

%\bibliographystyle{apalike}
\bibliography{BIB}
\bibliographystyle{acl_natbib}

\pagebreak
\appendix
\section{Model Architecture}\label{sec:supplementary}
%We first provide a detailed architecture of our two models, \uncond and \cond.
Once again, we wish to translate a source sequence $x_1^m \triangleq \langle x_1, \cdots, x_m \rangle$ into a target sequence $y_1^n \triangleq \langle y_1, \cdots, y_n \rangle$, and also predict image features $v$.

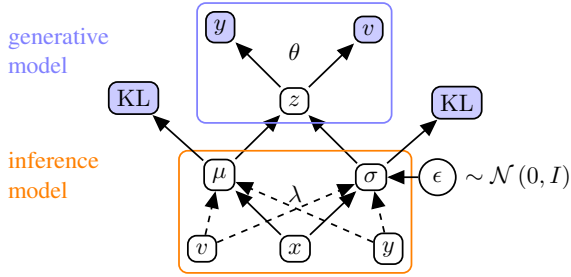
\begin{figure}[h!]
\center
\resizebox{\linewidth}{!}{
\begin{tikzpicture}[node distance=1cm]
\node[rectangle, draw, rounded corners, thick] (input) {$x$};
\node[rectangle, draw, rounded corners, thick, left=of input] (input_v) {$v$};
\node[rectangle, draw, rounded corners, thick, right=of input] (input_y) {$y$};
\node[rectangle, draw, rounded corners, thick, above left=of input] (mu) {$ \mu $};
\node[rectangle, draw, rounded corners, thick, above right=of input] (var) {$ \sigma $};
\node[rectangle, draw, rounded corners, thick, above right= of mu] (z) {$ z $};
\node[rectangle, fill=blue!20, thick, above left=of z, rounded corners, draw, node distance=1.5cm] (output) {$ y $};
\node[rectangle, fill=blue!20, thick, above right=of z, rounded corners, draw, node distance=1.5cm] (output_v) {$ v $};

\node[above =0.4cm of input] (lambda) {$ \lambda $};
%\draw[->, thick] (input) -- (mu) node[midway, above, rotate=315] {$ \lambda $};%{$ \lambda $};
\draw[->, thick] (input) -- (mu) node[midway, above, rotate=0] {$ $};%{$ \lambda $};
\draw[->, thick] (input) -- (var) node[midway, above, rotate=45] {$ $};
\draw[dashed, ->, thick] (input_v) -- (mu) node[midway, left, rotate=75] {$ $};
\draw[dashed, ->, thick] (input_v) -- (var) node[midway, above, rotate=45] {$ $};
\draw[dashed, ->, thick] (input_y) -- (mu) node[midway, above, rotate=315] {$ $};
\draw[dashed, ->, thick] (input_y) -- (var) node[midway, above, rotate=45] {$ $};
\draw[->, thick] (mu) edge (z);
\draw[->, thick] (var) edge (z);
\draw[->, thick] (z) -- (output) node[midway, above] {$  $}; %{$ \theta $};
\draw[->, thick] (z) -- (output_v) node[midway, above] {$  $}; %{$ \theta $};
%\node[above=of z, node distance=0.5cm] (theta) {$ \theta $};
\node[above =0.3cm of z] (theta) {$ \theta $};

\node[draw=orange, thick, rectangle, fit= (input) (input_v) (input_y) (mu) (var), rounded corners] {};
\node[left= of mu, text width=2cm] (inference) {\textcolor{orange}{inference model}};

\node[draw=blue!50, thick, rectangle, fit= (z) (output) (output_v), rounded corners] {};
\node[above= of inference, text width=2cm] (generation) {\textcolor{blue!50}{generative model}};

\node[circle, draw, thick ,right =of var, xshift=-.5cm] (epsilon) {$ \epsilon $};
\node[right = of epsilon, xshift=-1cm] (stdNormal) {$ \sim \NDist{0}{I} $};
\draw[->, thick] (epsilon) edge (var);

\node[above left= of mu, rectangle, draw, fill=blue!20, thick, rounded corners, thick] (KLmu) {$ \KullbackLeibler $};
\draw[->, thick] (mu) edge (KLmu);
\node[above right= of var, rectangle, draw, fill=blue!20, thick, rounded corners, thick] (KLvar) {$ \KullbackLeibler $};
\draw[->, thick] (var) edge (KLvar);
\end{tikzpicture}}
\caption{\label{fig:model_illustration}Illustration of multi-modal machine translation generative and inference models. The conditional model \cond includes dashed arrows; the fixed prior model \uncond does not, i.e. its inference network only uses $x$.}
\end{figure}

In Figure~\ref{fig:model_illustration}, we illustrate generative and inference networks for models \cond and \uncond.

\subsection{Generative model}
\paragraph{Source-language encoder}
The source-language encoder is deterministic and implemented using a 2-layer bidirectional Long Short-Term Memory (LSTM) network~\citep{HochreiterSchmidhuber1997_LSTM}:
\begin{equation}\label{eq:source_language_encoder}
\begin{aligned}
    \bm{f}_i &= \emb(x_i; \theta_\text{emb-x}), \\
    \bm{h}_0 &= \bm{\vec{0}}, \\
    \bm{\overrightarrow{h}}_i &= \LSTM(\bm{h}_{i-1}, \bm{f}_i; \theta_\text{lstmf-x}), \\
    \bm{\overleftarrow{h}}_i &= \LSTM(\bm{h}_{i+1}, \bm{f}_i; \theta_\text{lstmb-x}), \\
    \bm{h}_i &= [\bm{\overrightarrow{h}}_i, \bm{\overleftarrow{h}}_i],
\end{aligned}
\end{equation}
where $\emb$ is the source look-up matrix, trained jointly with the model, and $\bm{h}_1^m$ are the final source hidden states.

\paragraph{Target-language decoder}
Now we assume that $z$ is given, and will discuss how to compute it later on. % (we will address it for each model \uncond and \cond later on).
The translation model consists of a sequence of draws from a Categorical distribution over the target-language vocabulary (independently from image features $v$):
\begin{equation}
    Y_j | z, x, y_{<j} \sim \Cat(f_\theta(z, x, y_{<j} )), \notag
\end{equation}
where $f_\theta$ parameterises the distribution with an attentive encoder-decoder architecture:
%Specifically, our target-language decoder is a conditional LSTM:
\begin{align}
    \bm{w}_j &= \emb(y_j; \theta_\text{emb-y}), \notag\\
    \bm{s}_0 &= \tanh(\affine(\bm{h}_1^m; \theta_\text{init-y})), \notag\\
    \bm{s}_j &= \LSTM(\bm{s}_{j-1}, [\bm{w}_j, \bm{z}]; \theta_\text{lstm-y}), \notag\\
    \bm{c}_{i,j} &= \attention(\bm{h}^m_1, \bm{s}^n_{1}; \theta_\text{attn}), \notag\\
    f_\theta(z, x, y_{<j}) &= \softmax(\affine([\bm{s}_j, \bm{c}_j]; \theta_\text{out-y})), \notag
\end{align}
where the attention mechanism is a bilinear attention~\citep{Luongetal2015}, and the generative parameters are $\theta = \{ \theta_\text{emb-\{x,y\}}, \theta_\text{lstm\{f,b\}-x}, \theta_\text{init-y}, \theta_\text{lstm-y}, \theta_\text{attn}, \theta_\text{out-y} \}$.

\paragraph{Image decoder}
We do not model images directly, but instead as a $2048$-dimensional feature vector $\bm{v}$ of pre-activations of a ResNet-50's \texttt{pool5} layer.
We simply draw image features from a Gaussian observation model:
\begin{align}
	V|z &\sim \mathcal N(\boldsymbol \nu, \varsigma^2 I), \notag\\
    %V|z &\sim \mathcal N(\boldsymbol \nu, \diag(\boldsymbol \varsigma^2))  \\
    \boldsymbol \nu &= \MLP(z; \theta), \label{eq:mlp_image_decoder}%\\
    %\boldsymbol \varsigma &= f_\varsigma(z; \theta) ~ ,
\end{align}
where a multi-layer perceptron (MLP) %$f_\nu(\cdot)$ 
maps from $z$ to a vector of locations $\boldsymbol \nu \in \mathbb R^o$, and $\varsigma \in \mathbb R_{>0}$ is a hyper-parameter of the model (we use $1$).

\paragraph{Conditional prior \cond}
Given a source sentence $x_1^m$, we draw an embedding $z$ from a latent Gaussian model:
%\begin{equation}
\begin{align}
	Z|x_1^m &\sim \mathcal N(\boldsymbol\mu, \diag(\boldsymbol\sigma^2)), \notag\\
    \boldsymbol\mu &= \MLP(\bm{h}_1^m; \theta_\text{latent}), \label{eq:mlp-mu} \\
    \boldsymbol\sigma &= \softplus(\MLP(\bm{h}_1^m; \theta_\text{latent})) ~, \label{eq:mlp-sigma}
\end{align}
%\end{equation}
where Equations~(\ref{eq:mlp-mu}) %$f_\mu(\cdot)$ 
and~(\ref{eq:mlp-sigma}) % $f_\sigma(\cdot)$ 
employ two multi-layer perceptrons (MLPs) to map from a source sentence (i.e. source hidden states) to a vector of locations $\boldsymbol\mu \in \mathbb R^c$ and a vector of scales $\boldsymbol\sigma \in \mathbb R^c_{>0}$, respectively.

\paragraph{Fixed prior \uncond}
In the MMT model \uncond, we simply have a draw from a standard Normal prior:
\begin{equation}
	Z \sim \mathcal N(0, I). \notag\\
\end{equation}
%i.e. $\bm{z}$ is drawn from a standard normal prior.

%\paragraph{Multi-layer perceptrons} 
All MLPs have one hidden layer and are implemented as below (\cref{eq:mlp_image_decoder,eq:mlp-mu,eq:mlp-sigma}):
\begin{equation}
    \MLP(\cdot) = \affine(\ReLU(\affine(\: \cdot\: ; \theta)); \theta).\notag
\end{equation}

\subsection{Inference model}
The inference network shares the source-language encoder with the generative model and differs depending on the model (\cond or \uncond).

\paragraph{Conditional prior \cond}
Model \cond's approximate posterior $q_\lambda(z | x^m_1, y^n_1, v)$ is a Gaussian distribution:
\begin{equation*}
	Z | x_1^m, y_1^n, v \sim \mathcal N(\mathbf u, \diag(\mathbf s^2); \lambda).
\end{equation*}
We use two bidirectional LSTMs, one over source- and the other over target-language words, respectively. To reduce the number of model parameters, we re-use the entire source-language BiLSTM and the target-language embeddings in the generative model but prevent updates to the generative model's parameters by blocking gradients from being back-propagated (Equation~\ref{eq:source_language_encoder}).
Concretely, the inference model is parameterised as below:
\begin{align}
    \bm{h}_1^m &= \detach(\BiLSTM(x_1^m; \theta_\text{emb-x,lstmf-x,lstmb-x})), \notag\\
    \bm{w}_1^n &= \detach(\emb(y_1^n; \theta_\text{emb-y})), \notag\\
    \bm{h}_x &= \avg(\affine(\bm{h}_1^m; \lambda_\text{x})), \notag\\
    \bm{h}_y &= \avg(\BiLSTM(\bm{w}_1^n; \lambda_\text{y})), \notag\\
    \bm{h}_v &= \MLP(\bm{v}; \lambda_\text{v}), \notag\\
    \bm{h}_\text{all} &= [\bm{h}_x, \bm{h}_y, \bm{h}_v], \notag\\
    \mathbf u &= \MLP(\bm{h}_\text{all}; \lambda_\text{mu}), \notag\\
    \mathbf s &= \softplus(\MLP(\bm{h}_\text{all}; \lambda_\text{sigma})), \notag
\end{align}
where the set of the inference network parameters are $\lambda = \{ \lambda_\text{x}, \lambda_\text{y}, \lambda_\text{v}, \lambda_\text{mu}, \lambda_\text{sigma} \}$.

\paragraph{Fixed prior \uncond}
Model \uncond's approximate posterior $q_\lambda(z | x^m_1)$ is also a Gaussian:
\begin{equation*}
	Z | x_1^m \sim \mathcal N(\mathbf u, \diag(\mathbf s^2); \lambda),
\end{equation*}
where we re-use the source-language BiLSTM from the generative model but prevent updates to its parameters by blocking gradients from being back-propagated (Equation~\ref{eq:source_language_encoder}).
Concretely, the inference model is parameterised as below:
\begin{align}
    \bm{h}_1^m &= \detach(\BiLSTM(x_1^m; \theta_\text{emb-x,lstmf-x,lstmb-x})), \notag\\
    \bm{h}_x &= \avg(\affine(\bm{h}_1^m; \lambda_\text{x})), \notag\\
    \mathbf u &= \MLP(\bm{h}_\text{x}; \lambda_\text{mu}), \notag\\
    \mathbf s &= \softplus(\MLP(\bm{h}_\text{x}; \lambda_\text{sigma})), \notag
\end{align}
where the set of the inference network parameters are $\lambda = \{ \lambda_\text{x}, \lambda_\text{mu}, \lambda_\text{sigma} \}$.

Finally, all MLPs are implemented as below:
\begin{equation}
    \MLP(\cdot) = \affine(\ReLU(\affine(\: \cdot\: ; \lambda)); \lambda).\notag
\end{equation}
\end{document}